\documentclass[11pt]{article}

\usepackage[a4paper,margin=2.4cm]{geometry}
\usepackage{amsmath}
\usepackage{array}
\usepackage{booktabs}
\usepackage{caption}
\usepackage{float}
\usepackage{graphicx}
\usepackage{hyperref}
\usepackage{microtype} 
\usepackage[numbers,sort&compress]{natbib}
\usepackage{tabularx}
\usepackage[table]{xcolor}

\definecolor{ScientificDataHeader}{HTML}{F2AD72}
\definecolor{ScientificDataRule}{HTML}{777777}
\newcommand{\scientificdatatablestyle}{%
  \small%
  \renewcommand{\arraystretch}{1.15}%
  \arrayrulecolor{ScientificDataRule}%
}
\newcommand{\dataset}{4D-ImageNet}
\newcommand{\paperauthors}{%
  Yuyan Guan\textsuperscript{1},
  Haoran Zhang\textsuperscript{2},
  Zian Mao\textsuperscript{1},
  Antong Yang\textsuperscript{1},
  Caifei Li\textsuperscript{4},
  Jialong Wang\textsuperscript{4},
  Chuying Ouyang\textsuperscript{4},
  Hong Wang\textsuperscript{3,*},
  Xiaoqin Zeng\textsuperscript{3,*},
  Yujun Xie\textsuperscript{1,*}
}
\newcommand{\affiliationone}{Global Institute of Future Technology, Shanghai Jiao Tong University, Shanghai 200240, China}
\newcommand{\affiliationtwo}{Global College, Shanghai Jiao Tong University, Shanghai 200240, China}
\newcommand{\affiliationthree}{Materials Science and Engineering, Shanghai Jiao Tong University, Shanghai 200240, China}
\newcommand{\affiliationfour}{Contemporary Amperex Technology Co., Ltd., Ningde, P.R. China}
\newcommand{\correspondingauthors}{Hong Wang, Xiaoqin Zeng and Yujun Xie}
\newcommand{\correspondingemailone}{hongwang2@sjtu.edu.cn}
\newcommand{\correspondingemailtwo}{xqzeng@sjtu.edu.cn}
\newcommand{\correspondingemailthree}{yujun.xie@sjtu.edu.cn}
\newcommand{\code}[1]{\texttt{#1}}
\newcommand{\num}[1]{#1}
\newcommand{\SI}[2]{#1\,#2}
\newcommand{\angstrom}{\ensuremath{\text{\AA}}}
\newcolumntype{Y}{>{\raggedright\arraybackslash}X}
\hypersetup{hidelinks}
\renewenvironment{abstract}{%
  \section*{Abstract}\normalfont\normalsize\noindent
}{\par\vspace{1.0em}}

\begin{document}
{\noindent\LARGE\bfseries\raggedright
A benchmark dataset and baseline methods for four-dimensional STEM diffraction patterns\par}
\vspace{0.8em}
{\noindent\normalsize\raggedright
\paperauthors\par}
\vspace{0.65em}
{\noindent\small\raggedright
\textbf{Affiliations:}\par
\vspace{0.2em}
\textsuperscript{1}\affiliationone\par
\textsuperscript{2}\affiliationtwo\par
\textsuperscript{3}\affiliationthree\par
\textsuperscript{4}\affiliationfour\par
\vspace{0.35em}
\textsuperscript{*}\textbf{Corresponding authors:} \correspondingauthors. Email: 
\href{mailto:\correspondingemailone}{\correspondingemailone},
\href{mailto:\correspondingemailtwo}{\correspondingemailtwo}, and
\href{mailto:\correspondingemailthree}{\correspondingemailthree}\par}
\vspace{1.0em}
\begin{abstract}
Four-dimensional scanning transmission electron microscopy (4D-STEM) records a two-dimensional diffraction pattern at each electron-probe position, yielding spatially resolved reciprocal-space information but large, heterogeneous data volumes. Here we describe \dataset, a collection of 174,000 diffraction patterns comprising 145,000 experimental patterns selected from 29 acquisitions and 29,000 multislice simulations. The experimental data cover acquisition-level labels for Ag, Au, mixed Au--Ag, CoO, Pd and ZnO specimens across multiple fields of view, scan dimensions, camera lengths and exposure times. Each acquisition contributes 5,000 quality-ranked patterns with source scan coordinates and acquisition metadata. A set-prediction detector provides model-derived pseudo-labels for the direct-beam position and Bragg-disk centres, with a confidence score for each disk. The simulation data cover 13 crystal structures and include Euler rotations, reciprocal-space sampling and approximate low-index beam directions. A grouped mixed-domain masked-reconstruction benchmark is provided to assess leakage-resistant loading and evaluation across experimental and simulated data. The dataset is intended for representation learning, disk detection, diffraction-pattern retrieval, orientation analysis and simulation-to-experiment studies.
\end{abstract}

\section*{Background \& Summary}

In four-dimensional scanning transmission electron microscopy (4D-STEM), a focused electron probe is rastered over a specimen and a two-dimensional convergent-beam electron diffraction pattern is recorded at each two-dimensional probe position. The resulting four-dimensional datacube contains spatially resolved signatures of crystal structure, orientation, strain, electric and magnetic fields, and specimen thickness. These measurements connect real and reciprocal space, but their size and detector-specific formats complicate sharing, inspection and algorithm development\cite{ophus2019,savitzky2021}.

Machine-learning methods can use the detector plane as an image, learn compact representations across scan positions, or predict physically meaningful quantities from individual diffraction patterns. Progress is nevertheless limited by three recurring data problems. First, experimental 4D-STEM data are commonly retained as complete instrument datacubes whose size and proprietary or detector-specific layout hinder reuse. Second, diffraction intensities span a large dynamic range and contain detector artefacts, while the scientifically relevant diffracted disks may occupy only a small fraction of the detector. Third, adjacent scan positions are strongly correlated, so random pattern-level train--test splitting can overestimate generalisation. Dataset descriptors in related diffraction domains have shown the value of explicit simulation parameters, machine-readable labels and baseline tasks\cite{rincon2025}; equivalent documentation is needed for heterogeneous experimental 4D-STEM collections.

Here we describe \dataset, a diffraction-pattern collection organised around acquisition-level provenance and pattern-level metadata. The dataset contains two complementary components (Table~\ref{tab:overview}). The experimental component comprises 145,000 patterns selected from 29 4D-STEM acquisitions representing six specimen-level material groups. The simulation component comprises 29,000 patterns generated from crystallographic information files (CIFs) for 13 structures using abTEM\cite{madsen2021}. The two components are not paired; they provide complementary experimental diversity and simulation metadata, and Ag is the only material represented in both components.

Scan positions in each experimental acquisition were scored using the intensity-distribution criterion described below, and the 5,000 highest-ranked patterns were retained. The dataset preserves the original scan coordinates, rank, score and constituent score terms for every selected pattern, maintaining traceability to the source scan. The curated files are not complete dense 4D-STEM datacubes and are therefore unsuitable for quantitative virtual imaging or spatial-field reconstruction without an explicit missing-data treatment. The simulation files complement the experimental data with known Euler rotations, reciprocal-space pixel sampling, diffraction-centre coordinates and approximate low-index beam directions.

The experimental data combine acquisition-level material and microscope metadata with model-derived Bragg-disk pseudo-labels. A set-prediction network identifies candidate disk centres directly from each diffraction image and records a confidence score for every retained coordinate, providing machine-readable point annotations for supervised or confidence-weighted learning. A compact U-Net trained jointly on experimental and simulated patterns provides a masked-reconstruction usability benchmark evaluated on acquisition-held-out experimental data and file-held-out simulations. Together, the annotations and diffraction arrays enable disk detection, retrieval, weakly supervised material classification, orientation analysis and simulation-to-experiment studies.

\begin{table}[htbp]
\centering
\caption{Composition of \dataset. File sizes are decimal gigabytes measured for the PKL files.}
\label{tab:overview}
\small
\scientificdatatablestyle
\begin{tabularx}{\linewidth}{|Y|c|c|r|c|c|}
\hline
\rowcolor{ScientificDataHeader}
\textbf{Component} & \shortstack{\textbf{Material}\\\textbf{groups}} & \shortstack{\textbf{Files or}\\\textbf{acquisitions}} & \textbf{Patterns} & \shortstack{\textbf{Pattern}\\\textbf{shape}} & \shortstack{\textbf{Stored}\\\textbf{size}} \\ \hline
Experimental & 6 & 29 & \num{145,000} & $192\times192$ & \SI{21.39}{GB} \\ \hline
Simulation & 13 & 13 & \num{29,000} & $256\times256$ & \SI{7.60}{GB} \\ \hline
Total & 17 & 42 & \num{174,000} & -- & \SI{28.99}{GB} \\ \hline
\end{tabularx}
\end{table}

\begin{figure}[H]
\centering
\includegraphics[width=\linewidth]{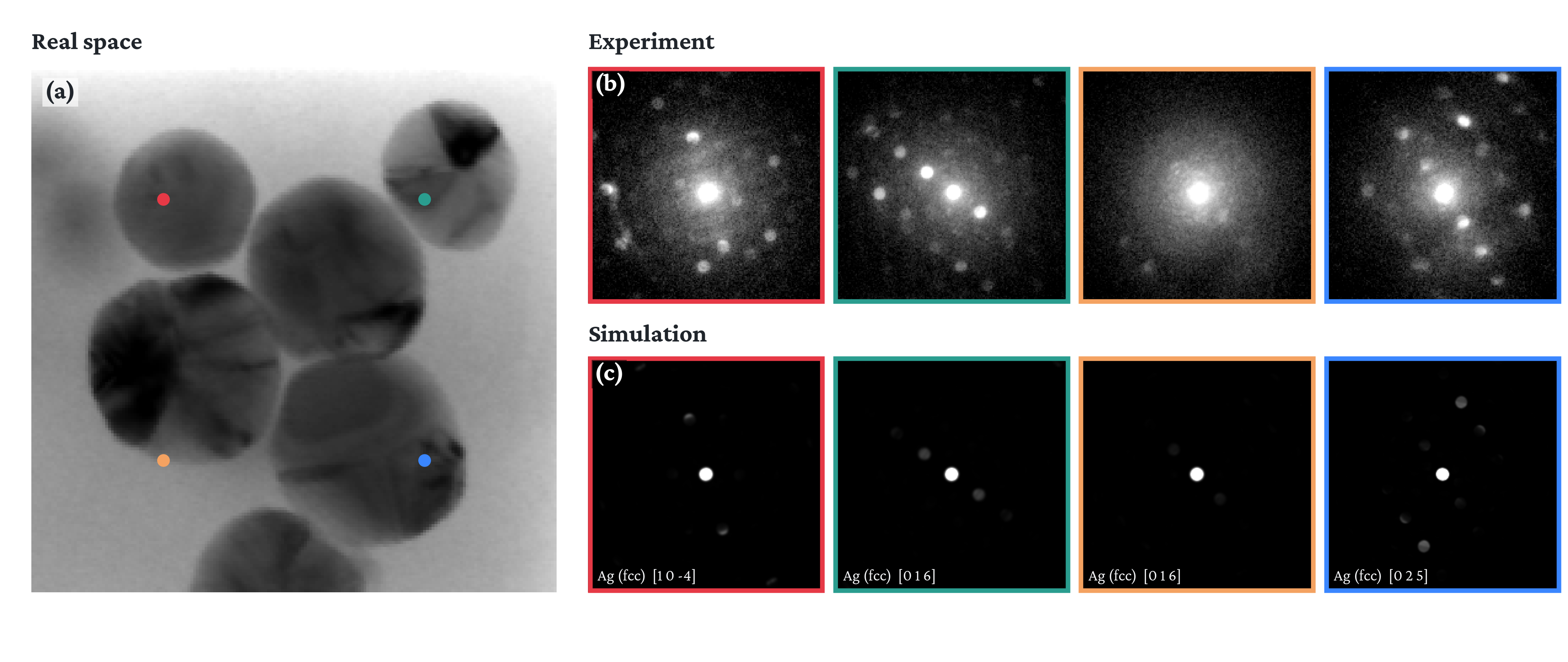}
\caption{Experimental and simulated 4D-STEM examples. (a) Experimental virtual bright-field image obtained by integrating a 12-pixel-radius region around the direct beam of an Ag acquisition; coloured markers identify four probe positions. Display used 0.5th--99.5th percentile clipping and linear normalisation. (b) Experimental diffraction patterns recorded at those positions, displayed after non-negative clipping, $\log(1+I)$ transformation, 15th--99.5th percentile clipping and a display gamma of 1.35. (c) Four distinct Ag multislice patterns selected by maximum cosine similarity of central-beam-excluded reciprocal-space features. The material phase and approximate $[hkl]$ beam direction are shown within each pattern; display used $\log(1+I)$, 15th--99.8th percentile clipping and a gamma of 0.92. Within each column, the border colour links the probe position, experimental frame and similarity-selected simulation. The simulations are visual references rather than paired ground truth, and all transformations are display-only.}
\label{fig:data-samples}
\end{figure}

Fig.~\ref{fig:data-samples} connects the real-space scan, experimental detector frames and complementary simulation component. The four experimental diffraction patterns are read from colour-coded probe positions in the same Ag acquisition. For each frame, a distinct pattern is selected from the Ag simulation component by maximising cosine similarity between central-beam-excluded reciprocal-space features. The selected simulations support visual comparison of reciprocal-space geometry but are neither paired reconstructions nor ground-truth counterparts of the experimental frames.

\section*{Methods}

\subsection*{Data Collection}

The experimental component comprises nanoparticle specimens labelled Ag, Au, mixed Au--Ag, CoO, Pd and ZnO. These names are specimen-level acquisition labels and should not be interpreted as local phase assignments or guarantees of phase purity. Nanoparticles with sizes of 20--\SI{100}{nm} were dispersed in ethanol or chloroform and ultrasonicated to disperse agglomerates. The resulting suspensions were drop-cast onto ultrathin carbon films supported on 300-mesh copper TEM grids.

Experimental data were acquired on a Thermo Fisher Scientific Talos F200X G2 microscope operated at an accelerating voltage of \SI{200}{kV} and equipped with a DECTRIS ARINA electron-counting detector. The convergence semi-angle recorded for all acquisitions was \SI{2.1}{mrad}. Detector frames in the curated files have $192\times192$ pixels. Acquisition fields of view were 95, 134, 190, 268 or \SI{379}{nm}; source scan arrays ranged from $128\times128$ to $260\times260$ probe positions. Nominal camera lengths were 98 or \SI{160}{mm}, and frame dwell or exposure times ranged from 1.5 to \SI{4}{ms}. A microscope spot-size setting of 7 is encoded in the source names of 16 acquisitions; the remaining 13 source names do not encode this setting, and no value is inferred for them.

The 29 acquisitions contain \num{1,737,302} source scan positions in total, calculated from the recorded scan dimensions. The source scans span six geometries (Table~\ref{tab:scan-shapes}) and multiple acquisition conditions. Repeated acquisitions with otherwise identical filename-level settings were retained as independent acquisition groups, preserving run-level variation and enabling complete acquisitions, rather than individual patterns, to be assigned to machine-learning partitions.

\begin{table}[htbp]
\centering
\caption{Experimental source-scan dimensions represented in the curated dataset.}
\label{tab:scan-shapes}
\scientificdatatablestyle
\begin{tabular}{|l|c|c|c|}
\hline
\rowcolor{ScientificDataHeader}
\textbf{Scan shape} & \shortstack{\textbf{Number of}\\\textbf{acquisitions}} & \shortstack{\textbf{Source positions}\\\textbf{per acquisition}} & \shortstack{\textbf{Retained patterns}\\\textbf{per acquisition}} \\ \hline
$128\times128$ & 1 & \num{16,384} & \num{5,000} \\ \hline
$130\times130$ & 1 & \num{16,900} & \num{5,000} \\ \hline
$185\times185$ & 2 & \num{34,225} & \num{5,000} \\ \hline
$200\times200$ & 1 & \num{40,000} & \num{5,000} \\ \hline
$256\times256$ & 13 & \num{65,536} & \num{5,000} \\ \hline
$260\times260$ & 11 & \num{67,600} & \num{5,000} \\ \hline
\end{tabular}
\end{table}

The material distribution is given in Table~\ref{tab:experimental-materials}. A fixed sample count of 5,000 patterns was retained from each acquisition rather than sampling in proportion to scan size. This prevents the largest scans from dominating the collection but means that the curated patterns are not a random sample of all probe positions.

\begin{table}[htbp]
\centering
\caption{Experimental material groups and numbers of acquisition-level files. Every acquisition contributes 5,000 selected patterns.}
\label{tab:experimental-materials}
\scientificdatatablestyle
\begin{tabular}{|l|r|r|}
\hline
\rowcolor{ScientificDataHeader}
\textbf{Material label} & \textbf{Acquisitions} & \textbf{Patterns} \\ \hline
Ag & 4 & \num{20,000} \\ \hline
Au & 3 & \num{15,000} \\ \hline
Au--Ag & 3 & \num{15,000} \\ \hline
CoO & 9 & \num{45,000} \\ \hline
Pd & 5 & \num{25,000} \\ \hline
ZnO & 5 & \num{25,000} \\ \hline
Total & 29 & \num{145,000} \\ \hline
\end{tabular}
\end{table}

\begin{figure}[H]
\centering
\includegraphics[width=\linewidth]{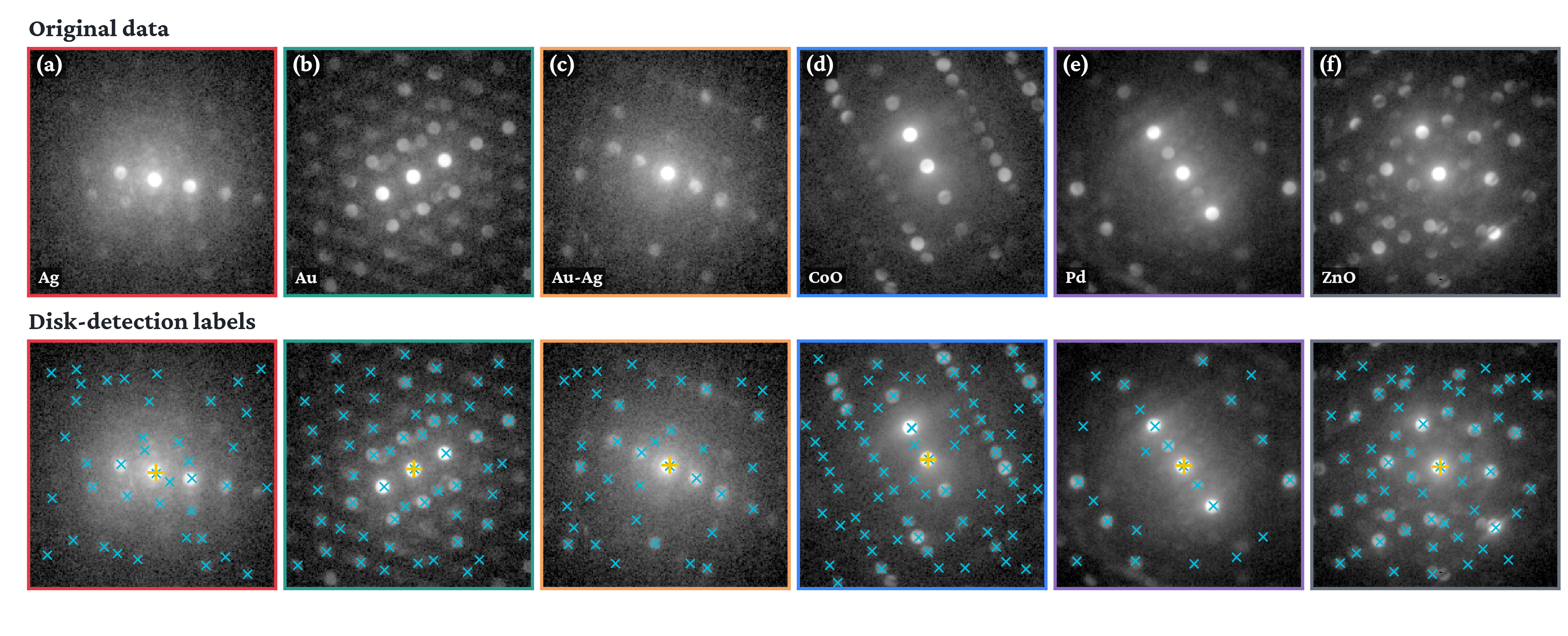}
\caption{Experimental diffraction patterns and corresponding model-derived pseudo-labels across the six material groups. Panels (a--f) show Ag, Au, Au--Ag, CoO, Pd and ZnO, respectively. The upper row contains the original experimental patterns and the lower row shows the identical detector pixels with r4 predictions read directly from the corresponding acquisition-level label JSON files at a confidence threshold of 0.5. Cyan crosses indicate predicted Bragg-disk centres, and the yellow plus sign indicates the output of the dedicated direct-beam query. For display, each pattern was independently transformed by non-negative clipping, $\log(1+I)$ and 1st--99.8th percentile normalisation. The overlays are point annotations rather than disk radii or segmentation masks; no diffraction feature was inserted, removed or redrawn.}
\label{fig:disk-label-examples}
\end{figure}

Fig.~\ref{fig:disk-label-examples} shows representative model-derived pseudo-labels for the experimental component. The six columns follow the material order used throughout the dataset, and each lower panel preserves the detector pixels of the panel immediately above it. Only point coordinates read from the corresponding acquisition-level label JSON files are added for visualisation. The set-prediction architecture used to generate these annotations is described under Construction of a Standardised Dataset and summarised in Fig.~\ref{fig:disk-architecture}.

\subsection*{Multislice simulation}

The simulation component was generated with abTEM\cite{madsen2021} from CIF structures obtained from the Materials Project\cite{jain2013} and read directly with the Atomic Simulation Environment\cite{larsen2017}. It contains Ag, CeO$_2$, Cu, CuO, Fe, GaAs, LiCoO$_2$, MoS$_2$, Pt, Si, SrTiO$_3$, Ti and TiO$_2$. Twelve files contain 2,000 patterns each and the TiO$_2$ file contains 5,000 patterns, giving 29,000 patterns in total.

For each structure, an orthogonalised supercell was expanded and cropped to a nominal lateral width of \SI{150}{\angstrom} and thickness of \SI{100}{\angstrom}. The potential sampling was \SI{0.1}{\angstrom}; one \SI{100}{\angstrom} potential slice was used. A \SI{200}{keV} probe with a \SI{2.1}{mrad} semi-angle cutoff was propagated to a pixelated detector with a maximum scattering angle of \SI{41.9}{mrad}. Thermal disorder was represented by eight frozen-phonon configurations. The isotropic displacement standard deviation was drawn independently and uniformly from 0 to \SI{0.1}{\angstrom} for each simulated pattern.

The specimen was rotated by independent Euler angles $\phi\sim U(0,360^\circ)$, $\theta\sim U(0,180^\circ)$ and $\psi\sim U(0,360^\circ)$. Rotations were applied about the specimen centre using \code{Atoms.euler\_rotate} in ASE, with angles in degrees and its passive $z$--$x$--$z$ convention. For column-vector coordinates relative to the rotation centre, the transformation is $R_z(-\psi)R_x(-\theta)R_z(-\phi)$, where $R_x$ and $R_z$ denote right-handed active rotation matrices. The stored triplets describe this operation on the orthogonalised, CIF-derived supercell. Independent uniform sampling of these angles does not give a uniform distribution over three-dimensional orientations.

The simulated diffraction array was centred by an affine transformation, resampled to $256\times256$ pixels, and convolved with a Gaussian kernel whose standard deviation was drawn uniformly from 0.5 to 1.5 pixels. Each final simulated array was divided by its total intensity to produce a non-negative, unit-sum intensity array. No Poisson counting noise, detector point-spread function, inelastic background, scan distortion or experimental dead-pixel pattern was added. These omissions should be considered when transferring a model trained only on the simulations to experimental data.

The stored orientation is the applied Euler-angle triplet in degrees. An accompanying integer $[h,k,l]$ label approximates the crystallographic plane normal parallel to the incident beam. It was obtained by transforming the laboratory beam direction into the unrotated crystal frame and searching primitive reciprocal-lattice vectors with indices from $-6$ to 6 for the maximum absolute directional cosine. The sign was canonicalised so that the first non-zero index is positive. The $[h,k,l]$ value is therefore a nearest low-index direction under a finite search, not a refined zone-axis determination. The reciprocal-space pixel increments and diffraction centre calculated by abTEM are retained for every pattern.

\subsection*{Construction of a Standardised Dataset}

The collection was organised at the acquisition level to preserve provenance and support leakage-resistant machine-learning splits. One merged experimental PKL file corresponds to one source acquisition and contains exactly 5,000 selected patterns; one simulation PKL file corresponds to one material structure and contains all patterns generated from that structure. Experimental and simulated arrays are stored as 32-bit floating-point values. Automated disk labels are stored separately and linked to an experimental pattern by acquisition identifier and array index. Label generation does not change the associated diffraction intensities, simulated diffraction features are not inserted into experimental patterns, and simulation orientation labels are not assigned to experimental data.

For each source acquisition, a sparse average diffraction pattern was calculated from at most 256 approximately evenly spaced scan positions. The beam centre $\mathbf{c}_0=(c_y,c_x)$ was estimated as the intensity centre of mass of this average. Three masks were then defined on the detector: a central disk of radius 12 pixels around $\mathbf{c}_0$, an outer region at radii of at least 20 pixels, and the complementary non-central region.

Every diffraction pattern $I$ was evaluated without changing its stored intensity values. Let $T=\sum I$ be its total intensity, $f_c$ the fraction of $T$ inside the 12-pixel central mask, $f_o$ the fraction in the outer region, $d$ the Euclidean distance in pixels between the pattern centre of mass and $\mathbf{c}_0$, $\sigma_{nc}$ the standard deviation in the non-central region, and $\bar I$ the pattern mean. The bounded or scaled terms were

\begin{align}
B_c &= \max\left(0,1-\frac{|f_c-0.22|}{0.18}\right),\\
B_a &= \max\left(0,1-\frac{d}{12}\right),\\
B_v &= \frac{\min\left(\sigma_{nc}/\max(\bar I,10^{-6}),10\right)}{10}.
\end{align}

The final selection score was

\begin{equation}
S = 0.35B_c + 0.35f_o + 0.20B_a + 0.10B_v.
\end{equation}

Patterns with zero or invalid total intensity receive non-informative metric values and are not favoured by this ranking. A bounded min-heap retained the 5,000 highest-scoring scan positions per acquisition. Selected patterns were first written as individual PKL records containing source identity, two-dimensional scan coordinate, score, rank, estimated beam centre and component metrics. The records were then sorted by rank and merged into one acquisition-level PKL file. The merge operation stacks arrays and metadata without changing detector intensities.

The ranking retains patterns with measurable off-axis diffraction signal while avoiding severe direct-beam displacement or domination. It is a curation heuristic rather than a physical ground-truth label and changes the empirical intensity and spatial distributions relative to the complete datacube. The score and its component terms are retained so that this selection effect can be analysed or filtered.

Pattern-level disk coordinates were generated with the r4 SupervisedDNSR detector. The model treats the disks in one diffraction pattern as an unordered set of points rather than as a dense segmentation mask. A ConvNeXt V2 Tiny backbone\cite{woo2023} extracts four feature levels, which are projected to a common token dimension of 192. Four multi-scale deformable-attention encoder layers\cite{zhu2021} integrate local high-resolution responses with broader diffraction context. Encoder objectness and coordinate heads propose reference points, after which 384 learned disk queries and one separate direct-beam query are refined by six decoder layers. Each disk query predicts a disk or no-object logit and a normalised two-dimensional coordinate; the direct-beam query predicts one normalised coordinate independently of the disk set.

The pretrained r4 checkpoint was held fixed and used in evaluation mode to annotate \dataset; no parameter updates were performed during this annotation workflow. The accompanying supervised-training implementation uses separate loaders for simulated and experimental diffraction records, with reference disk-centre coordinates stored in \code{qi\_list} (row) and \code{qj\_list} (column). Supervision is point-based rather than pixel-wise segmentation. Hungarian bipartite matching\cite{carion2020} associates predicted and reference coordinates using classification and $L_1$ coordinate costs. The objective combines disk classification, matched-coordinate regression and direct-beam regression with weights 2, 5 and 1, respectively, and applies auxiliary losses to the decoder stages. The supplied training configuration specifies 150 epochs, AdamW with an initial learning rate of $10^{-4}$, five warm-up epochs and a cosine learning-rate schedule. The released annotations are model-derived pseudo-labels, not independently established ground truth.

For application to \dataset, each diffraction pattern was clipped to non-negative values, transformed with $\log(1+I)$, divided by its maximum and bilinearly resized to $256\times256$ pixels. The final decoder produced normalised coordinates $(x_n,y_n)\in[0,1]^2$ and disk probabilities. Predictions with confidence greater than 0.5 were retained and mapped back to the original detector dimensions as $x=x_nW$ and $y=y_nH$. The direct-beam output is a learned regression target and is distinct from the intensity centre of mass used during pattern curation. A disk coordinate denotes the predicted centre of a diffraction disk; the detector does not predict disk radius or a segmentation boundary.

One acquisition-level JSON annotation file is generated for each experimental PKL file. Its top-level fields record the schema version, source file, material, acquisition identifier, file-level beam-centre estimate and model configuration. The \code{predictions} array contains one object per diffraction pattern with the pattern index, source scan position, original image shape, predicted direct-beam coordinate, disk count and a variable-length list of disk coordinates and confidence scores. Pixel and normalised coordinates are stored together, maintaining traceability between annotations and diffraction arrays.

\begin{figure}[H]
\centering
\includegraphics[width=\linewidth]{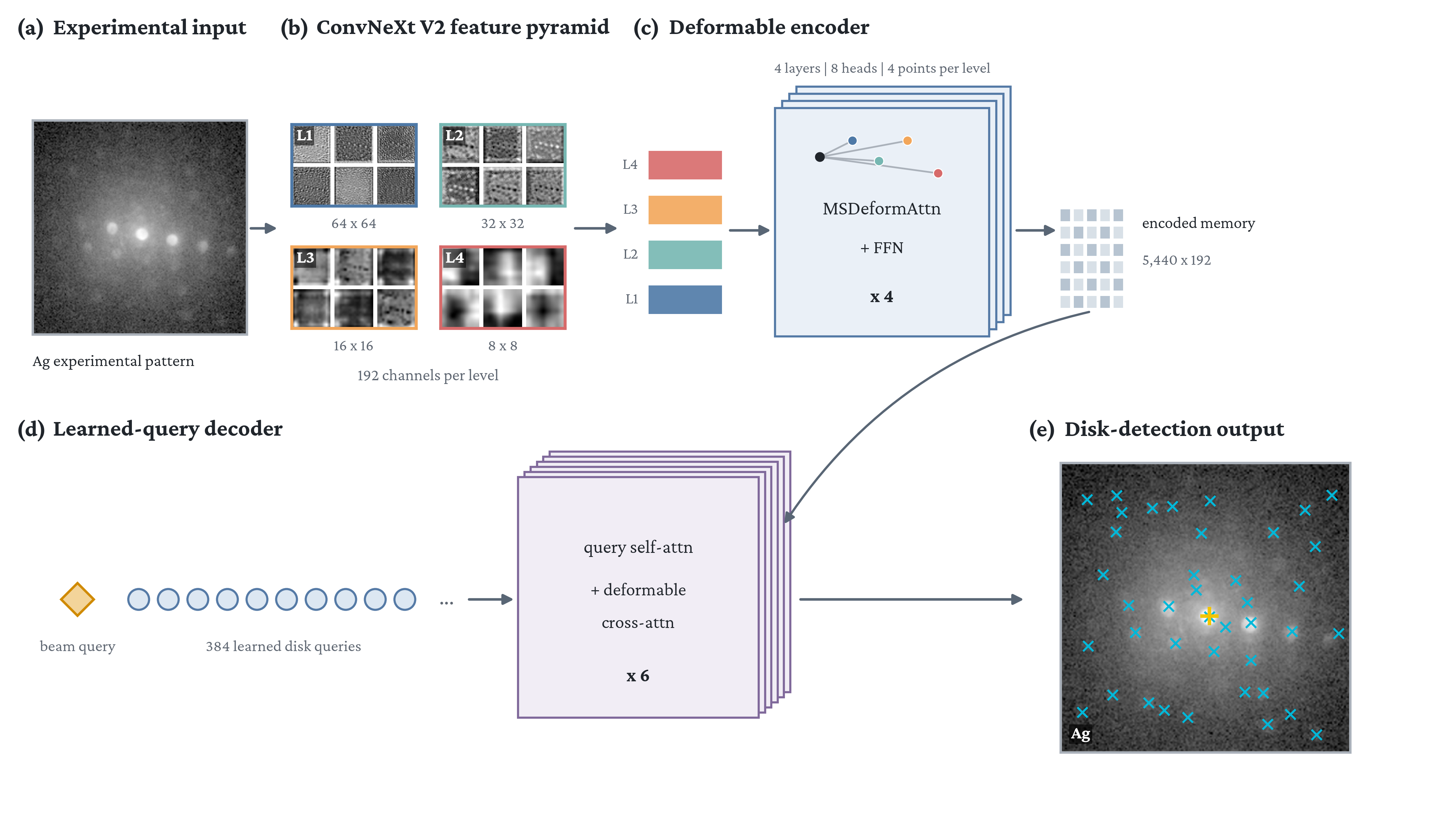}
\caption{Schematic diagram of the disk-detection annotation architecture. (a) A real experimental Ag diffraction pattern is used as input. (b) A ConvNeXt V2 Tiny backbone produces four feature levels of $64\times64$, $32\times32$, $16\times16$ and $8\times8$, each projected to 192 channels. (c) Four multi-scale deformable-attention encoder layers, each using eight heads and four sampling points per feature level, form a $5{,}440\times192$ memory. (d) Six decoder layers refine one dedicated beam query and 384 learned disk queries through self-attention and deformable cross-attention. (e) Predictions at a confidence threshold of 0.5 are overlaid on the unchanged input: cyan crosses denote predicted Bragg-disk centres and the yellow plus sign denotes the predicted direct-beam position. The feature panels are stored activations from the frozen r4 checkpoint, and the markers are read from the corresponding acquisition-level label JSON file.}
\label{fig:disk-architecture}
\end{figure}

Fig.~\ref{fig:disk-architecture} summarises the complete annotation path from an experimental detector frame to the predicted direct-beam coordinate and set of Bragg-disk centres. The displayed feature maps are stored activations from the frozen r4 checkpoint, and the output panel reads the prediction for the same Ag input from its acquisition-level label JSON file at a confidence threshold of 0.5.

\section*{Data Records}

The \dataset\ data record is available from Science Data Bank (ScienceDB)\cite{guan2026dataset} and comprises 29 acquisition-level experimental PKL files, 13 structure-level simulation PKL files, 29 acquisition-level disk-annotation JSON files and a JSON manifest. Table~\ref{tab:records} summarises the records required for reuse.

\begin{table}[htbp]
\small
\centering
\caption{Core data and annotation records in \dataset.}
\label{tab:records}
\scientificdatatablestyle
\begin{tabularx}{\linewidth}{|>{\raggedright\arraybackslash}p{0.20\linewidth}|Y|}
\hline
\rowcolor{ScientificDataHeader}
\textbf{Record} & \textbf{Contents} \\ \hline
Experimental PKL & One acquisition per file; 5,000 $192\times192$ patterns in \code{diffraction\_patterns}, with \code{material}, \code{source\_name}, \code{scan\_positions}, \code{scan\_shape}, \code{fov}, \code{camera\_length} and selection \code{scores}. \\ \hline
Disk-annotation JSON & One acquisition per file, linked to its PKL file by \code{source\_file}. The \code{predictions} array contains one object per pattern; \code{center\_beam}, \code{disk\_count} and \code{disks} provide direct-beam and disk-centre coordinates. The top-level \code{model} object records the detector version, weight identifier, preprocessing, input size and confidence threshold. \\ \hline
Label manifest JSON & One dataset-level file listing the source and annotation path for every acquisition together with pattern- and disk-count summaries. \\ \hline
Simulation PKL & One structure per file; $256\times256$ patterns with \code{material}, \code{source\_cif}, \code{center\_list}, \code{pixel\_size\_list}, \code{orientation\_list}, \code{hkl\_list} and simulation parameters. \\ \hline
\end{tabularx}
\end{table}

In each object of the \code{predictions} array, \code{center\_beam} contains $(x,y)$ in original detector pixels and the corresponding normalised unit-square coordinate. The \code{disks} list contains the same coordinate representations and a confidence score for each predicted disk centre. The list length equals \code{disk\_count}; it varies among patterns and may be zero. These are model-derived point annotations rather than disk radii, pixel masks or independently verified ground truth. The detector version, weight identifier and threshold are retained so that annotations can be regenerated or filtered.

Experimental material labels apply at acquisition level, whereas disk coordinates apply to individual patterns. Simulation orientations and $[h,k,l]$ values follow from the simulation procedure and are not assigned to experimental images. No calibrated reciprocal-space pixel coordinate is stored for the experimental patterns; the nominal camera length alone is insufficient for quantitative conversion from detector pixels to scattering vector.

\section*{Data Overview}

\begin{figure}[H]
\centering
\includegraphics[width=0.96\linewidth]{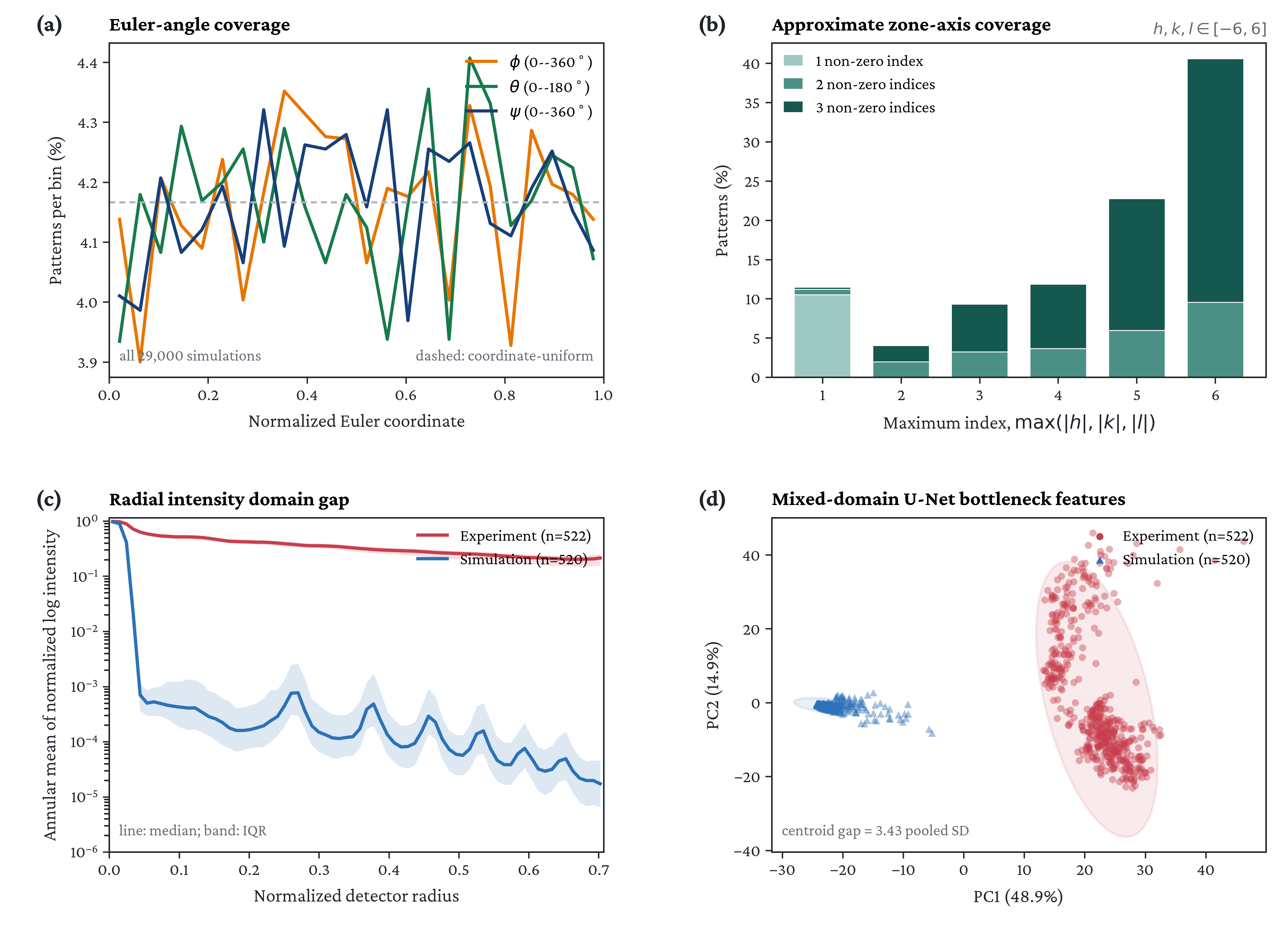}
\caption{Simulation coverage and experiment--simulation domain gap. (a) Marginal distributions of the three Euler coordinates for all 29,000 simulations after normalisation by their respective sampling ranges; the dashed line denotes the expectation for a coordinate-uniform distribution. (b) Approximate $[h,k,l]$ coverage grouped by the maximum absolute index and the number of non-zero components. These low-index directions are simulation metadata and are not crystallographically equivalent phase labels across materials. (c) Radial profiles for 522 experimental and 520 simulated patterns processed with the same intensity transform and normalised detector radius. At each radius, the line is the median across patterns of the per-pattern annular mean of normalised log intensity, and the band spans the interquartile range. (d) Joint principal-component projection of frozen bottleneck features from the best mixed-domain U-Net checkpoint. Both domains were forwarded as unmasked $[x,0]$ inputs; shaded ellipses show 90\% covariance contours. The centroid separation of 3.43 pooled within-domain spread units documents residual domain structure and is not an accuracy score.}
\label{fig:simulation-domain-gap}
\end{figure}

The arrays span six experimental material groups and 13 simulated structures, with acquisition and simulation metadata retained at file level. The experimental conditions are not factorially balanced: CoO contributes nine acquisitions, whereas Au and Au--Ag each contribute three, and fields of view, camera lengths and exposure times are associated with particular acquisition groups. Consequently, models may exploit acquisition-specific contrast or detector signatures as shortcuts for specimen-level material labels. Fig.~\ref{fig:simulation-domain-gap} summarises simulation coverage and residual experiment--simulation differences. Its Euler-angle and approximate low-index-direction panels use all 29,000 simulations; the radial-intensity comparison uses 522 experimental and 520 simulated patterns processed identically. Simulated patterns exhibit lower median diffuse off-axis intensity than experimental patterns. In the frozen bottleneck feature space of the jointly trained masked U-Net, the first two joint principal components explain 48.9\% and 14.9\% of the variance, and the domain centroids differ by 3.43 pooled within-domain spread units. These statistics describe dataset composition and residual domain structure; they do not measure reconstruction or domain-adaptation accuracy.

\section*{Technical Validation}

\subsection*{Interpretation of automated disk labels}

The disk coordinates are generated by a trained detector and are therefore provided as pseudo-labels rather than independent ground truth. Fig.~\ref{fig:disk-label-examples} verifies coordinate alignment between the diffraction arrays and acquisition-level JSON records, while Fig.~\ref{fig:disk-architecture} documents the model path that generated them; neither figure constitutes an independent accuracy assessment of the r4 checkpoint. The confidence values are ranking scores from the disk/no-object softmax and should not be interpreted as calibrated probabilities. Downstream evaluation of disk detection should use an independently annotated reference set and should report precision, recall, localisation error and direct-beam error separately.

The pseudo-labels remain useful as reproducible point targets for representation learning, retrieval and detector pre-training. Confidence thresholds can be increased for applications that favour precision, while continuous scores can be retained as sample weights. Ambiguous, overlapping, edge-truncated and weak disks require particular care because the stored coordinates encode centres only and do not describe disk extent or uncertainty.

\subsection*{Masked-reconstruction usability benchmark}

A mixed-domain masked-reconstruction experiment was used to verify joint indexing, acquisition- or file-level splitting, batching and model evaluation for the experimental and simulation components. A compact two-level U-Net\cite{ronneberger2015} received two co-registered channels: the diffraction pattern with selected off-axis disk regions masked and the corresponding binary mask. Each encoder stage applies two $3\times3$ Conv--BN--ReLU layers; $2\times2$ max pooling reduces the feature-map resolution from $192\times192$ to $96\times96$ and then $48\times48$, while the channel count increases from 16 to 32 and 64. Each decoder stage begins with a $2\times2$ transposed convolution, concatenates the corresponding encoder feature map and applies two further Conv--BN--ReLU layers. A $1\times1$ convolution followed by a sigmoid produces the one-channel prediction. Before masking, each pattern was converted to non-negative 32-bit floating point, transformed using $\log(1+I)$, divided by its maximum and resized to $192\times192$ pixels when necessary; experimental patterns therefore retained their native spatial dimensions, whereas simulations were resized from $256\times256$.

Candidate off-axis intensity maxima were detected outside a central exclusion radius of 0.06 times the image dimension. Between one and four peak-centred circular regions were sampled per pattern, with a disk radius of 0.035 times the image dimension, an 11-pixel peak window and a minimum relative peak intensity of 0.12. The model was trained to reconstruct the masked intensities with an intensity-weighted masked mean-squared error.

The benchmark sampled 5,000 experimental and 5,000 simulated patterns. Experimental acquisitions were split by acquisition identifier while retaining every material in each partition; simulation files were split as complete groups. The mixed training set contained 6,390 patterns (2,929 experimental patterns from 17 acquisitions and 3,461 simulations from nine files), and validation contained 1,806 patterns (1,036 experimental patterns from six acquisitions and 770 simulations from two files). Independent tests contained 1,035 experimental patterns from six held-out acquisitions spanning all six experimental materials and 769 simulations from the held-out CeO$_2$ and SrTiO$_3$ files. Training used Adam with a learning rate of $10^{-3}$, batch size 8, automatic mixed precision and a maximum of 60 epochs. A validation-loss scheduler reduced the learning rate; the checkpoint with the lowest validation loss was obtained at epoch 48. Table~\ref{tab:baseline} reports both independent tests. Structural similarity was not computed and is therefore not reported.

\begin{table}[htbp]
\centering
\caption{Held-out masked-reconstruction metrics for the mixed-domain usability benchmark. Metrics are calculated on masked pixels after the stated preprocessing.}
\label{tab:baseline}
\scientificdatatablestyle
\begin{tabular}{|l|r|r|}
\hline
\rowcolor{ScientificDataHeader}
\textbf{Metric} & \textbf{Experimental test} & \textbf{Simulation test} \\ \hline
Number of patterns & 1,035 & 769 \\ \hline
Masked mean-squared error & 0.001851 & 0.002020 \\ \hline
Masked mean-absolute error & 0.03180 & 0.02273 \\ \hline
Intensity-weighted masked mean-squared error & 0.001717 & 0.003831 \\ \hline
Peak signal-to-noise ratio & \SI{27.33}{dB} & \SI{26.95}{dB} \\ \hline
\end{tabular}
\end{table}

\begin{figure}[t!]
\centering
\includegraphics[width=0.98\linewidth]{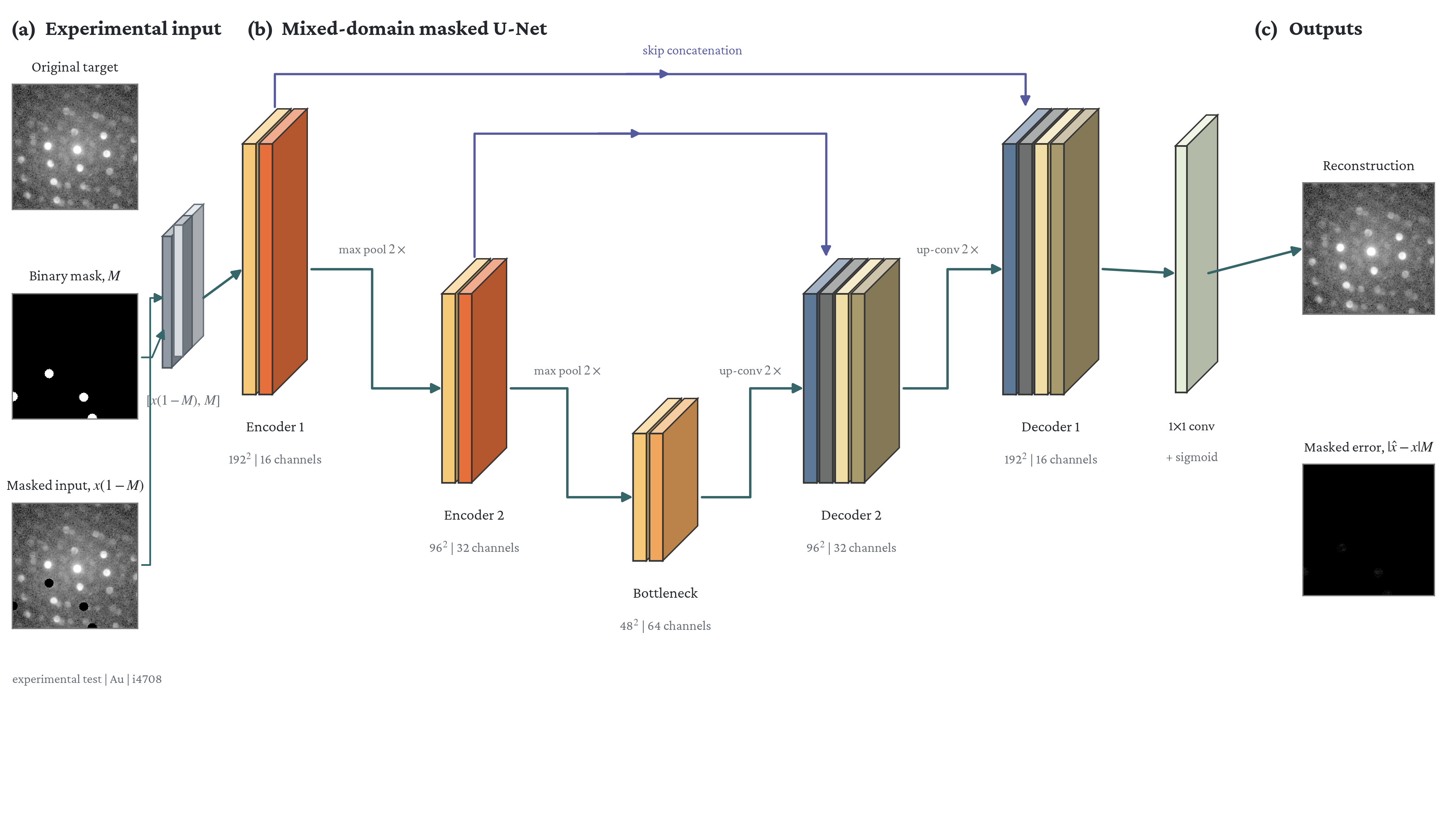}
\caption{Mixed-domain masked-reconstruction U-Net and experimental test example. (a) A held-out Au target, four-region binary mask $M$ and masked pattern $x(1-M)$ form the two-channel model input $[x(1-M),M]$. (b) The implemented Small U-Net is shown in a stepped encoder--decoder layout. Pairs of feature slabs represent the two $3\times3$ Conv--BN--ReLU layers in each encoder or bottleneck block. The feature maps progress from $192^2$ pixels and 16 channels to $96^2$ pixels and 32 channels, and then to the $48^2$-pixel, 64-channel bottleneck. Each decoder group shows the $2\times2$ transposed convolution, concatenated encoder feature and two decoder convolutions; purple routes denote the two skip concatenations. A $1\times1$ convolution and sigmoid produce the one-channel prediction. (c) The prediction is inserted into the masked regions to form the reconstruction, and the masked absolute error is displayed on the same target-intensity scale. Pattern 4,708 was drawn from a held-out experimental acquisition and evaluated using the mixed-domain epoch-48 checkpoint.}
\label{fig:masked-unet-architecture}
\end{figure}

These values verify end-to-end mixed-domain loading, grouped splitting, training and evaluation under the stated protocol. They do not establish physically exact recovery of unmeasured intensities and should not be compared across studies unless masking, preprocessing and grouping rules are held fixed. Fig.~\ref{fig:masked-unet-architecture} summarises the implemented architecture and shows a deterministic inference result for Au pattern 4,708 from an acquisition excluded from training and validation. The four masked off-axis regions, reconstruction and error were generated with the frozen epoch-48 checkpoint; no diffraction feature was drawn or inserted for the figure.

\section*{Usage Notes}

\subsection*{Loading and intensity preprocessing}

The PKL records can be read with a pinned Python environment containing a compatible NumPy version. The following minimal example reads one merged file and verifies alignment:

\noindent\begin{minipage}{\linewidth}
\begin{verbatim}
from pathlib import Path
import pickle
import numpy as np

path = Path("exp/Ag/<acquisition>/<acquisition>.pkl")
with path.open("rb") as handle:
    record = pickle.load(handle)

patterns = np.asarray(record["diffraction_patterns"], dtype=np.float32)
positions = np.asarray(record["scan_positions"], dtype=np.int32)
assert patterns.shape[0] == positions.shape[0] == record["count"]
\end{verbatim}
\end{minipage}

Pickle can execute code during loading and must only be opened from a trusted source. Pickles created with different NumPy major versions can refer to different internal module paths; users should load the files with the software environment specified alongside the processing code and, when broader interoperability is required, export the arrays once to a non-executable format such as HDF5 or Zarr.

Experimental and simulated intensities have deliberately different native scales. Experimental arrays retain their stored detector-intensity scale, whereas every simulated pattern is normalised to unit sum. For visualisation or joint model input, a reproducible transform such as non-negative clipping, $\log(1+I)$ and per-pattern maximum normalisation can reduce dynamic-range differences. The original stored arrays should be retained for analyses requiring quantitative intensities. Any clipping, centring, resizing or normalisation must be documented with model results.

\subsection*{Recommended split strategies}

Randomly splitting individual patterns from the same acquisition creates leakage because neighbouring probe positions share specimen structure and acquisition-specific detector conditions. Experimental splits should therefore use \code{source\_name} or another acquisition identifier as the minimum grouping unit. For stricter domain-generalisation tests, complete materials or acquisition-condition combinations may be held out. Simulation splits should group by \code{source\_cif} when testing structural transfer and by material when testing performance on unseen compositions. The baseline code defaults to one PKL file per group if no external metadata table is supplied.

Task-specific label interpretation is essential. The experimental \code{material} field is an acquisition-level label and is suitable for weakly supervised classification, retrieval or representation analysis, but it is not a pixel-level phase map. The simulation \code{orientation\_list} records the applied transform; \code{hkl\_list} is an approximate nearest low-index direction. Neither should be transferred to an experimental pattern as ground truth solely through visual similarity.

\subsection*{Spatial analyses and selection bias}

The selected scan coordinates allow users to map a prediction back to its source scan grid, but only at retained positions. Missing positions are not zero-intensity measurements and should be represented as unobserved. Dense virtual detector images, centre-of-mass field maps, ptychographic reconstruction and quantitative strain mapping generally require the complete datacube and should not be performed on the curated subset without an explicit missing-data method and validation against full scans.

Because the curation score favours central-beam balance, outer-region intensity, beam alignment and non-central variation, the dataset is enriched for patterns with discernible diffraction signal. It is not suitable for estimating the natural frequency of empty, amorphous, weak-scattering or severely miscentred frames in the source experiments. Analyses of anomaly prevalence or unbiased scan statistics require the complete source datacubes rather than the curated subsets.

\subsection*{Using the disk labels}

The automated disk coordinates provide model-derived pattern-level pseudo-labels for the experimental component. They can be used as point targets for detector pre-training, confidence-weighted representation learning, indexing of local reciprocal-space features or construction of disk-centred image patches. Users may apply a stricter confidence threshold or retain the continuous score as a sample weight; the threshold and detector version should always be reported. Because the annotations were generated by r4, they are not independent ground truth for evaluating the same checkpoint. Accuracy evaluation requires a separately expert-annotated reference set that was not used to train r4.

The direct-beam coordinate and Bragg-disk coordinates have different roles and should not be merged into one disk list. Coordinates are stored in $(x,y)$ order, while legacy training arrays use \code{qj} for $x$ and \code{qi} for $y$. Pixel coordinates refer to the original $192\times192$ experimental pattern, and normalised coordinates allow use after resizing. The present labels contain centre points and confidence values only; methods requiring disk radius, shape or segmentation must derive and validate those quantities separately.

\section*{Data Availability}

The \dataset\ dataset is available from Science Data Bank (ScienceDB)\cite{guan2026dataset} at {\urlstyle{same}\url{https://doi.org/10.57760/sciencedb.nbsdc.00281}}. Experimental and simulated diffraction patterns and their metadata are provided in PKL format, while acquisition-level Bragg-disk annotations and the dataset manifest are provided in JSON format. Details of the dataset contents, file organisation and variable definitions are provided in the Data Records section. The experimental component contains curated pattern subsets rather than complete dense 4D-STEM datacubes.

\section*{Code Availability}

The custom Python code used for experimental data preprocessing, multislice simulation, automated Bragg-disk annotation and masked-reconstruction benchmarking is publicly available at {\urlstyle{same}\url{https://github.com/Gaiya69-rgb/4D-ImageNet}}. The repository contains the model implementations, training and inference scripts, configuration files, dependency specifications and example data. The trained r4 disk-detector weights and the mixed-domain epoch-48 masked-reconstruction U-Net weights are available with the dataset in ScienceDB\cite{guan2026dataset} at {\urlstyle{same}\url{https://doi.org/10.57760/sciencedb.nbsdc.00281}}. The repository README specifies the local paths for the downloaded checkpoints and provides instructions for installation, execution and verification.

\section*{Author Contributions}

Y.G. and Y.X. conceived and designed the study. Y.G. prepared the specimens and performed the experimental 4D-STEM acquisitions. H.Z. conducted the multislice simulations. Y.G., J.W. and C.L. developed the standardised data-processing pipeline and constructed the 4D-STEM dataset. Y.G. and Z.M. developed and implemented the disk-detection model and generated the Bragg-disk annotations. Y.G. and A.Y. developed the masked-reconstruction U-Net and performed the representation-learning benchmarks using experimental and simulated diffraction patterns. C.O. and Y.X. provided expertise in electron microscopy, diffraction analysis and materials science. X.Z. and H.W. supervised the research and contributed to the interpretation of the results. Y.G. and Y.X. drafted the paper. All authors reviewed and approved the final paper.

\section*{Competing Interests}

The authors declare no competing interests.


\section*{Funding}

This work was supported by the National Natural Science Foundation of China (grant nos. 12474186, 52401159 and 52425101).

\section*{Ethics Statement}

This study describes inorganic materials data and did not involve human participants, human data or animals.

\end{document}